\documentclass[letterpaper, 10 pt, conference]{ieeeconf}  

\IEEEoverridecommandlockouts                              

\usepackage{graphics} 
\usepackage{epsfig} 
\usepackage{mathptmx} 
\usepackage{times} 
\usepackage{amsmath} 
\usepackage{amssymb}  
\usepackage{xcolor}
\usepackage{comment}
\usepackage{mwe}
\usepackage{caption}
\usepackage{graphicx}
\usepackage{pbox}
\usepackage{tabularx,booktabs}
\usepackage{float} 
\usepackage{subfigure}
\usepackage{comment}
\usepackage{soul}
\usepackage{gensymb}

\usepackage{fancyhdr}
\fancypagestyle{firstpage}{%
    \chead{This paper has been accepted for publication at the 2023 International Conference on Robotics and Automation.}
    }

\title{
A Novel Concentric Tube Steerable Drilling  Robot for Minimally Invasive Treatment of Spinal Tumors Using Cavity and U-shape Drilling Techniques
}

\author{Susheela Sharma$^{1}$, Ji H. Park$^{1}$, Jordan P. Amadio$^{2}$, Mohsen Khadem$^{3}$, and Farshid Alambeigi$^{1}$,\IEEEmembership{~Member,~IEEE}
\thanks{**This work is supported by the National Institute Of Biomedical Imaging and Bioengineering of the National Institutes of Health under Award Number R21EB030796.}
\thanks{$^{1}$S.~Sharma, J.~H.~Park, and F.~Alambeigi are with the Walker Department of Mechanical Engineering and the Texas Robotics  at the University of Texas at Austin, Austin, TX, 78712, USA. Email: \{sheela.sharma, jihwanpark98\}@utexas.edu,  farshid.alambeigi@austin.utexas.edu}.
\thanks{$^{2}$J.~P.~ Amadio is with the Department of Neurosurgery, The University of Texas Dell Medical School, TX, 78712. }
\thanks{$^{3}$M.~Khadem is with the School of Informatics, University of Edinburgh, UK.}
}

\begin{document}

\maketitle
\thispagestyle{firstpage}
\pagestyle{empty}


\begin{abstract}
In this paper, we present the design, fabrication, and evaluation of a novel flexible, yet structurally strong, Concentric Tube Steerable Drilling Robot (CT-SDR) to improve minimally invasive treatment of spinal tumors. Inspired by concentric tube robots, the proposed two degree-of-freedom (DoF) CT-SDR, for the first time,  not only allows a surgeon to intuitively and quickly  drill smooth  planar and out-of-plane J- and U- shape curved trajectories, but it also, enables drilling  cavities through a hard tissue in a minimally invasive fashion. We successfully evaluated the performance and efficacy of the proposed CT-SDR in drilling various planar and out-of-plane J-shape branch, U-shape, and cavity drilling scenarios on simulated bone materials. 
\end{abstract}

	\section{Introduction}
\begin{figure}[!t] 
		\centering 
		\includegraphics[width=0.7\linewidth]{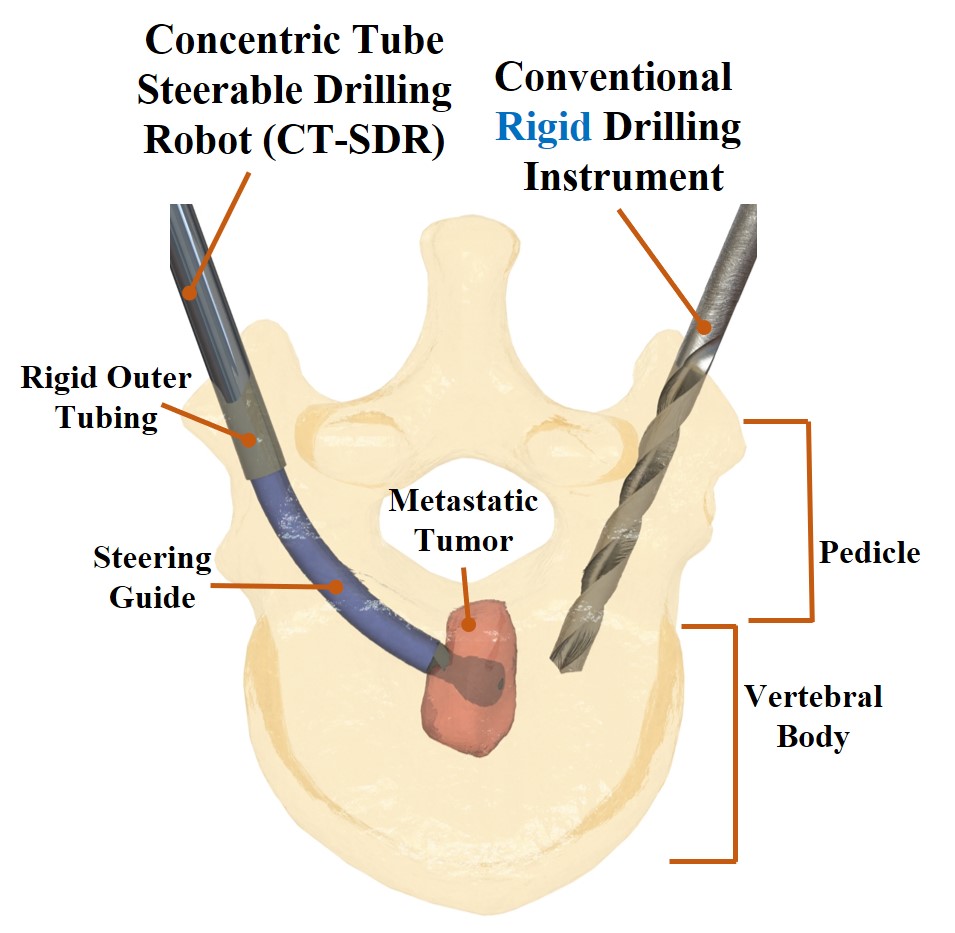}
		\caption{Conceptual illustration of the proposed CT-SDR, drilling to access a metastatic tumor in the vertebral body of  an L5 vertebra compared with a conventional rigid drilling instrument constrained to linear trajectories. As shown the proposed CT-SDR is composed of a stationary rigid outer tubing and a  steering guide for a drill tip.}
		\label{fig:concept}
	\end{figure}

	\vspace{-5 pt}
Bone is the most common site of metastatic disease after lung and liver \cite{Hatrick, Maccauro} and one of the most common causes of chronic pain among cancer patients \cite{Hatrick,Maccauro}. Each year, 400,000 people in the US alone suffer from bone metastases, which includes two-thirds of patients with metastatic disease \cite{Jacobs}. The most frequent site of bone metastasis is seen in the spine and particularly vertebrae, which includes 50\% of all bone metastatic disease (Fig. \ref{fig:concept}) \cite{Choi}. 
Traditional therapeutic methods for spinal tumors  involve chemotherapy and possibly radiation, pain management, etc. However, many patients have a minimal or only brief response to these therapies and onset of pain relief can take months \cite{anchala2014treatment}. Additionally, invasive surgical procedures depend on the nature and location of the tumor ({Fig. \ref{fig:concept}}) and are usually not warranted in these patient populations due to significant blood loss, post-op pain, wound healing issues, and coexisting health problems \cite{pusceddu2013treatment}. 

Despite invasive surgical methods, minimally invasive (MI) procedures such as thermal ablation, vertebroplasty, and kyphoplasty are safe and effective treatments of painful spinal metastatic lesions  \cite{steinmetz2001management}. In these manual image guided procedures, typically, a surgeon advances a rigid instrument through the skin and vertebra on a patient’s back under X-ray guidance to confirm that it has entered the lesion area in vertebra. Then, to stabilize the typically fractured or degraded bone in vertebra, a rigid pedicle screw is utilized and/or bone cement, using a rigid syringe, is injected into the fractured vertebra for fixation \cite{becker2008assessment}. However, due to the rigidity of the utilized instruments and complex and sensitive anatomy of spine surrounded by nerves, these procedures typically suffer from the lack of enough accessibility to the tumor lesion and therefore cannot completely remove/treat the tumor and in some cases  may even increase the risk of tumor spread to blood vessels \cite{nussbaum2004review}. 


    As shown in Fig. \ref{fig:concept}, to address the aforementioned limitation of existing rigid instruments, surgeons can use novel flexible robotic  systems to minimally invasively navigate to harder to reach regions within the vertebral body by drilling in curved trajectories and reach the tumor area (e.g., \cite{alambeigi2017curved,alambeigi2018inroads,alambeigi2019use, alambeigi2020steerable}). After the drilling  procedure, using the robotic system, different  treatment procedures can then be delivered locally and precisely to the tumor area to perform either a high dose radiation for brachytherapy  \cite{ZUCKERMAN2018e235} and ablation \cite{brace2009radiofrequency} procedures, or the robot can be used to completely excise the  tumor area (e.g., \cite{CochlearImplants, YANG2018103}). Nevertheless, these applications require a flexible, yet structurally strong, robot capable of safely drilling through the bone to access these difficult to reach areas without buckling or experiencing structural failure. Ensuring this balance between structural stiffness and flexible dexterity is the essential challenge in designing steerable drilling robotic systems \cite{alambeigi2019use,FlexibleReview,alambeigi2016toward}.  
    
    To overcome the aforementioned challenges, literature documents a few cases of flexible drilling robots which have been developed to improve access to areas within a patient's hard tissues. For instance, Alambeigi et al. \cite{alambeigi2017curved,alambeigi2019use} utilized a one degree-of-freedom (DoF) tendon-driven flexible manipulator to develop a robotic system to drill curved tunnels for treatment of femoral head osteonecrosis. However, the proposed one DoF system suffers from several limitations including a small workspace limited to only planar and short J-shape trajectories (only reaching 35 mm in length and 40\degree from the original cutting angle), and very slow drilling procedure (as long as 5-9 minutes). This can mainly be attributed to the restrictions placed on the geometry of the robot and tendon's maximum load capacity and the stiffness of the robotic system in regards to the force required to actuate a bend in a system of this size. This system was recently improved by Ma et al. \cite{ma2021active} to a handheld device while addressing certain system limitations. While the system could produce greater motion from the drill tip, similar to the previous robotic version, the stiffness and bending behavior of the robot are limited to the stiffness of the continuum manipulator and the max loading capacity of the actuation tendons. Of note, in these systems, the curvature of the drilled trajectory  is heavily reliant on the properties of the hard tissue being actively interacted with at the time of drilling, and cannot directly be controlled using the proposed force-based control procedure. Additionally, the steering of the proposed handheld device (i.e., simultaneous control of insertion and bending Degrees-of-Freedom (DoF)) is highly unintuitive and is greatly limited in its ability to produce accurate surgical procedures by the user's expertise. In other words, the user does not have a direct control on the curvature of the drilled trajectory and requires an active sensing approach to ensure the safety and accuracy of the procedure.
    In an effort towards addressing the limitations discussed previously, recently, Wang et al. \cite{wang2021design,9732206} have proposed an actively controlled tendon-driven surgical drilling system for  spinal applications. While this drill grants more access to areas within the hard tissue when compared to conventional rigid drilling instruments, it does not allow for smooth curved trajectories. In other words, the drill trajectories are restricted to multi-segment straight/linear paths due to the use of a rigid shaft with articulated wrist/hinge.
    
    To collectively address the aforementioned limitations and as our main contributions, in this paper, we propose the design, fabrication, and evaluation of a novel flexible yet structurally strong concentric tube steerable drilling robot (CT-SDR) to improve the process of spinal tumor treatments. Inspired by concentric tube robots (e.g., \cite{dupont2009design,webster2010design}), the proposed two DoF CT-SDR, for the first time,  not only allows the surgeon to  intuitively and quickly drill  planar and out-of-plane J- and U-shape curved trajectories,  but it also, enables drilling continuous cavities to completely excise an spinal tumor in a minimally invasive fashion. 
    Thanks to the utilized pre-shaped NiTi tubes, the robotic system also removes the unintuitive active control from the user and decouples the remaining DoFs to provide an easy-to-steer system.  This feature completely addresses the mentioned challenge in active steering of the previous robotic systems (i.e., \cite{alambeigi2017curved, ma2021active, wang2021design}). The performance of the proposed CT-SDR has been assessed under varying experimental conditions and goals while drilling both smooth curved trajectories and cavities within simulated bone materials.

 \section{Design and Fabrication of the CT-SDR} \label{sec:II}
	\vspace{-5 pt}
	To meet the needs of surgeons for complete treatment of spinal tumors, the  CT-SDR needs to provide  (i)  required DoFs to enable a planar and out-of-plane generic J- and U-shape drilling trajectories as well as enabling cavity cutting based on the geometry of the tumor, (ii)  flexible power transmission from a high rpm drill motor to carry rotational motion to the drill's cutting tip; (iii) sufficiently strong and flexible guides to steer the drill's cutting tip towards the areas of interest within the patient without deviation; and (iv) an actuation unit and control system to allow a surgeon to actively control the drill tip's position throughout a surgical procedure. The following sections will address in detail these different requirements and the components that were designed and manufactured to meet them \cite{sheela-ismr}.

	
	\begin{figure}[!t] 
		\centering 
		\includegraphics[width=0.7\linewidth]{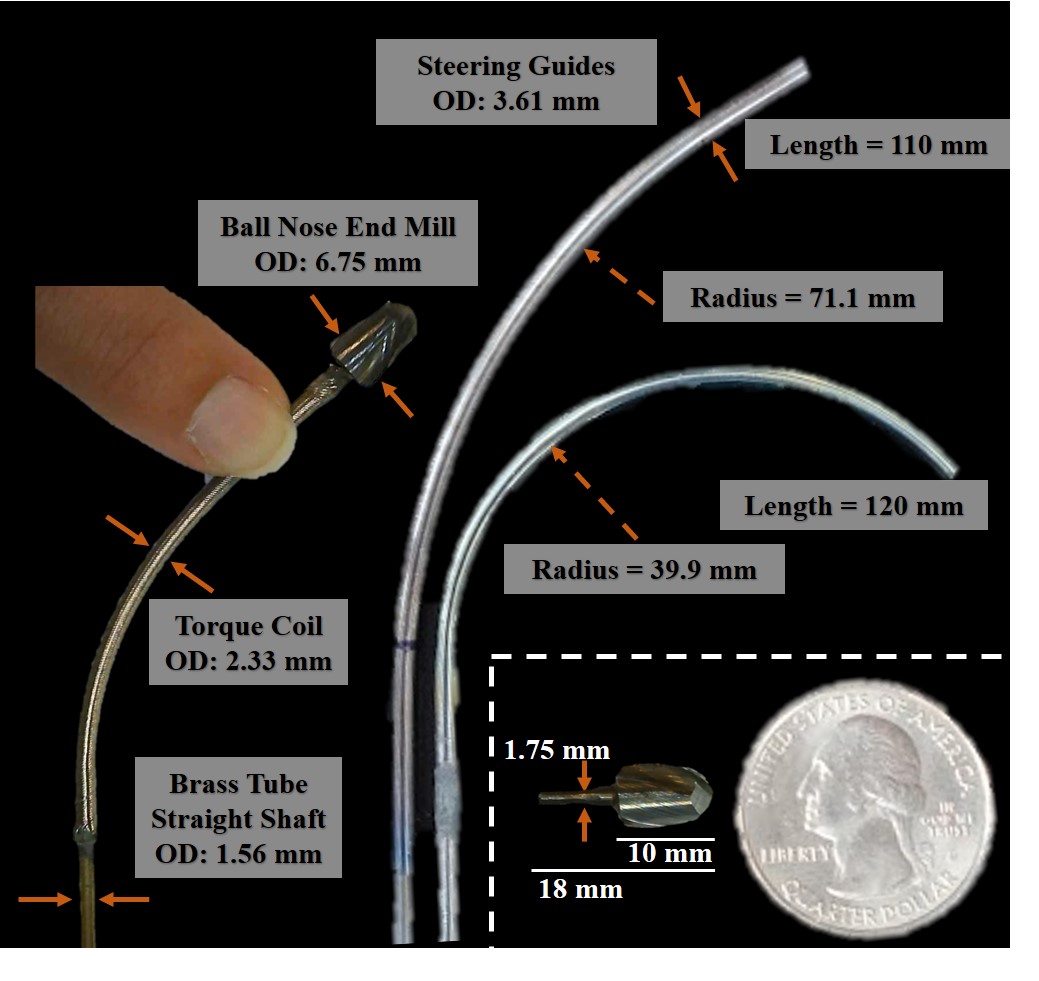}
		\caption{The NiTi steering guides and the flexible shaft that correspond to the two inner tube systems of the CT-SDR. A close view of the designed drill bit is shown in the subfigure.}
		\label{fig:flex}
	\end{figure}
	
		\subsection{Concentric Tube Steering Guides}
				\begin{figure*}[!t]
		\centering
	   \includegraphics[width=1\linewidth]{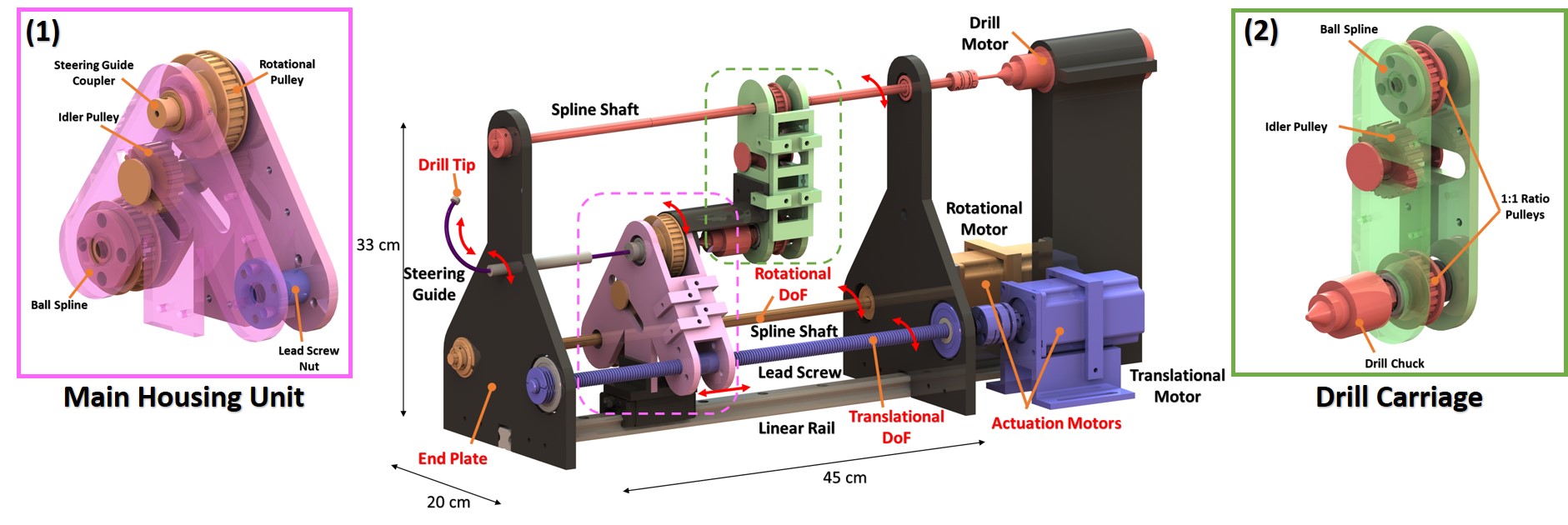}
		\caption{The design of the CT-SDR system. The rotational actuation sub-system is highlighted in orange, the translational subs-system in purple. (1) shows an interior view of the Main Housing Unit and the components of the rotational and translational sub-systems housed there. (2) Depicts an interior view of the drill carriage which houses a majority of the components to control the drill tip's rotational speed.}
	\label{fig:CAD}
	\end{figure*}
	\vspace{-5 pt}
	To create the required pathways and cavities, the CT-SDR required steering guides that could move the drill's cutting tip into areas of interest by the surgeon. These guides would need to be both flexible enough to bend outward from the drill's entry point to access hard-to-reach areas, but strong enough to not deflect under the forces experienced by the cutting tip during drilling. Our design takes advantage of the superelastic properties of NiTi metal (Euroflex GmbH, Germany), to provide a solution to these contrasting requirements. This superelastic, biocompatible, shape memory alloy is heat treated to a pre-designed curvature, which establishes the CT-SDR's range of motion. Following the heat treatment procedures provided in previous studies (e.g., \cite{hodgson2001fabrication}), the NiTi tubes in their original straight state were constrained to a desired shape using a CNC-fabricated stainless steel jig and placed in a furnace to create the designed drill trajectories. After heat treatment, the tubes had curvatures of 71.1 and 39.9 mm radii.  The steering guides used in this paper are shown in Fig. \ref{fig:flex}. Of note the selected curvatures and tubing dimensions were arbitrarily chosen based on the geometry of an L4 vertebra. Nevertheless, these curvatures can readily be changed depending on the vertebral level and geometry.

	When assembled into the CT-SDR system, the NiTi steering guide is nested within a larger stainless steel tube which provides the structural strength and rigidity required to constrain the NiTi tube into a straight configuration. This utilization of nestled tubes with differing rigidity pulls inspiration from designs of concentric tube robots for soft tissue manipulation (e.g., \cite{dupont2009design, webster2010design, burgner2015continuum}). The stainless steel tube, which holds the role of the concentric tube's outer tube and is shown in Fig. \ref{fig:concept}, is static in this design of the CT-SDR, and the NiTi steering guide is actuated through it. As the guide is moved forward and out of the constraining outer tube, the portion of the guide removed from the stainless steel returns to its heat treated, pre-programmed shape/curvature. In the process of returning to it's pre-programmed shape, the guide steers the drill's cutting tip along the guide's trajectory, creating a curved and smooth drilled path.

	\subsection{Flexible Power Transmission and Cutting Tool}\label{cutting tool}
	\vspace{-5 pt}
	As shown in Fig. \ref{fig:flex}, each flexible tool comprises of a small rigid cutting tip, a flexible torque coil, and a straight rigid tube. To secure the components to one another, epoxy (1813A243, McMaster-Carr) is used at the part intersections. 
	A ball nose end mill (8878A42, McMaster-Carr) was selected for the drill tip, as it produced clean smooth tunnels in earlier testing and has large flutes for faster material removal. Notably, the main concern in selecting a drill tip was the cutter's ability to remove material not only at the distal tip but also on the sides of the cutter during planar and out-of-plane drilling procedures.
	The cutting tip has a diameter of 6.75 mm, a cutting tip  10 mm in length, and a shank 8 mm in length with a ground down diameter to 1.75 mm. 
	The drill tip geometries and torque coil connection are shown in Fig. \ref{fig:flex}.
		The power transmission and the tool's flexibility was possible through the utilization of a torque coil (Asahi Intec. USA, Inc.) placed behind the drill's cutting tip. This torque coil is 115 mm in length, runs through the curved section of the NiTi tubing to serve as a method for delivering rotational motion around a curve in a reliable way. The coil did not connect directly to the drill chuck in the CT-SDR's design to avoid crush damage to the coil, and instead was attached to a straight brass tube (8859K231, McMaster-Carr), with a diameter of 1.56 mm. 

	\subsection{CT-SDR Actuation Design and Controls}
	\vspace{-5 pt}
	A primary requirement for the design of the two DoF CT-SDR, shown in Fig. \ref{fig:CAD}, was to keep the utilized motors that actuate the steering guides and the cutting tool stationary to minimize the inertia of the system's moving components and subsequently the required power for actuating the system. To satisfy this design requirement, we utilized different methods for transmitting both translational and rotational motions through the system, to produce the desired motion for the steering guides and the CT-SDR's drill tip.
			\begin{figure*}[!t]
		\centering
	   \includegraphics[width=0.8\linewidth]{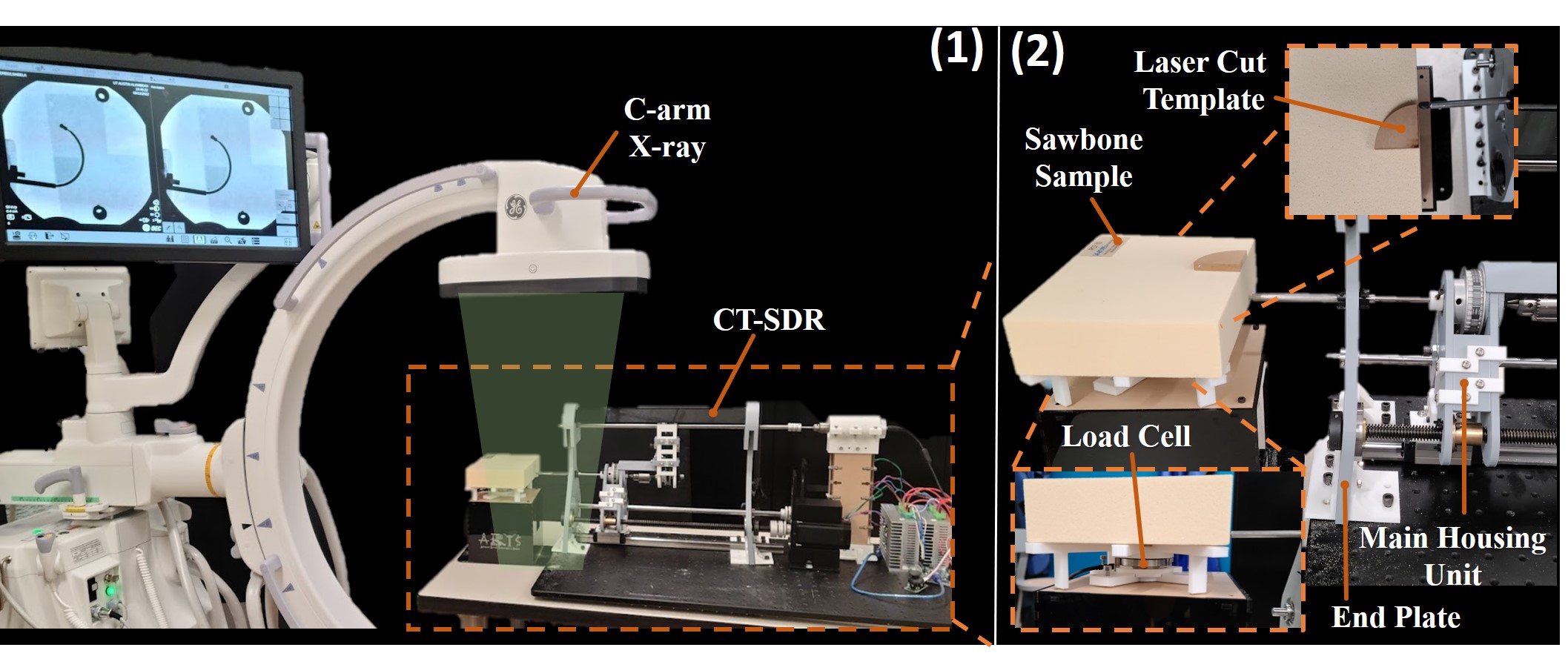}
		\caption{Experimental set-up used to evaluate the CT-SDR, including C-arm X-ray machine, a six DoF load cell, laser cut template and sawbone test sample. (1) An overview of the entire set-up with the C-arm's visual cone. (2) a closer view of the sample set up, and CT-SDR, including a side view of the load cell, and a top view of the alignment of the laser guide with the system. The Main Housing Unit of the CT-SDR can also be seen in this view.}
		\label{fig:set-up}
	\end{figure*}
	As shown in Fig. \ref{fig:CAD}, the CT-SDR's insertion DoF is controlled by a NEMA 23 stepper motor (6627T530, McMaster-Carr) rotating a lead screw (98940A305, McMaster-Carr) to adjust the position of a nut (6350K41, McMaster-Carr) rigidly held within the main housing unit. The main housing is supported by a carriage sliding on a linear rail (6709K431, McMaster-Carr), which allows for lower friction during translation as the lead screw actuates the housing. The NiTi steering guide's rotational orientation DoF is also controlled by a NEMA 23 stepper motor, this time controlling a spline shaft (61145K145, McMaster-Carr) which allows for the housing to have unrestricted motion along the linear rail, while still transmitting the rotational position of the connected stepper motor. The ball spline (61145K430, McMaster-Carr), within the main housing, is secured within a belt and pulley system connected to the NiTi steering guide's coupler. The designed pulleys were selected to have a 1:1 ratio for easy control by the stepper motor. An idler pulley was also designed into the system to ensure enough belt tension is maintained in the system. 

	To control the drill tip's rotational speed, a carriage was rigidly attached to the top of the main housing unit to serve as a channel for the high speed rotations of the drill motor to be transmitted through. The drill motor (B075SZZN4J, Amazon) is mounted in a custom holder above the stepper motors at the back of the system and connected to another spline shaft (61145K143, McMaster-Carr) that runs the length of the system. Similar to the steering guide's rotational control shaft, this one allows the carriages to slide freely along the shaft direction while transmitting the rotational motion provided by the drill motor. Another 1:1 pulley system with idler pulley transmits this motion to a drill chuck (2812A19, McMaster-Carr) mounted on a stainless steel hollow shaft. 
		Also, as shown in Fig. \ref{fig:CAD}, the NiTi steering guide is attached to the main housing unit with a designed 3D printed coupler and set screw, which allows for the main housing components to control the guide's position and orientation. End plates placed at either end of the system provide support for many of the actuation unit's moving parts. These plates and the stepper motor mounts were 3D printed in PLA and secured to an optical breadboard for stability. 

	The utilized stepper motors were controlled with Rtelligent R60 motor drivers (B07SBFZ596, Amazon), an Arduino Uno R3 microcontroller board, and a custom program written with the AccelStepper.h Arduino library. The written program allowed for independent control of both the insertion and rotation degrees of freedom, or could be modified to control these freedoms simultaneously. The speed of the system in both rotation and insertion were also adjustable allowing us to optimize the different settings we used in drilling.

	\section{Evaluation Experiments}
	\vspace{-5 pt}
	\subsection{Experimental Set-Up}
	\vspace{-5 pt}
 Fig. \ref{fig:set-up} shows the experimental setup used to thoroughly evaluate the pefrormance of developed CT-SDR in drilling planar and out-of-plane J- and U-shape trajectories together with creating cavities within a hard tissue.   The experiments used Sawbone biomechanical bone model phantoms (block 5 and 10 PCF, Pacific Research Laboratories, USA) to simulate diseased human bones with lower bone mineral densities compared with a healthy tissue \cite{ccetin2021experimental}. As can be seen in  Fig. \ref{fig:set-up}, the CT-SDR system, mounted on an optical breadboard, was placed on a wooden table with the specimen held in front of the drilling tip on an acrylic stand. The materials of the table and the stand were selected so that they showed minimal interference with the C-arm X-ray (OEC One CFD, GE Healthcare) placed next to the system to monitor the CT-SDR's progress through the test sample during experiments. The addition of the C-arm allowed for real time monitoring of the experiment by the user, and as an option for analysis after a test's conclusion. Views from the C-arm for different drilling experiments can be seen in Fig. \ref{fig:X-ray Progress}. To measure and compare trends in drilling forces on the CT-SDR's drill tip to previous studies, a six DOF force/torque load cell (Mini45, ATI Industrial Automation) was placed with a sample holder beneath the test specimen. An additional camera was placed separately from the other components to provide another viewpoint of the tests, and provide tracking for the experiments. Recorded videos of the performed experiments with these visuals have been provided in the complementary uploaded media file.
	
	\subsection{Drilling Experiments}
	\vspace{-5 pt}
	Each experiment run with the CT-SDR was designed to test the capabilities of the proposed two DoF robotic system and evaluate if it could reliably  produce 
	predictable long planar and out-of-plane curved drilling trajectories
	and, as shown in Fig. \ref{fig:concept}, minimally invasively remove cavities of material  by entering from a small hole (e.g., vertebrae's pedicle) with the diameter of the drill bit and remove materials within the anatomy (ie.e, vertebral body). 

	\subsubsection{U-shape Planar Drilling}\label{U-shape}
	
\begin{figure}[!t] 
		\centering 
		\includegraphics[width=0.6\linewidth]{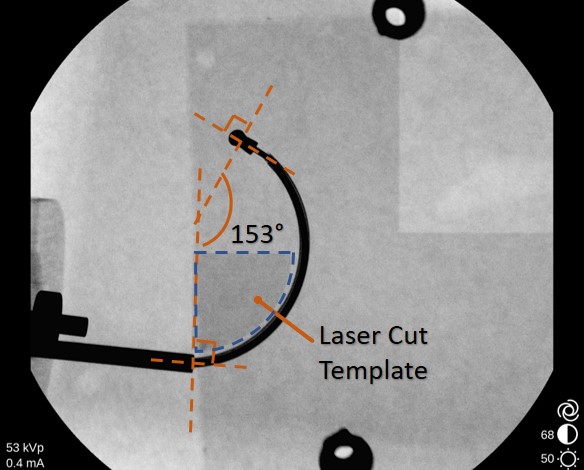}
		\caption{X-ray view of a U-shape trajectory test performed with a 39.9 mm steering guide in PCF 10 sawbone. Visible at the top is a 35 mm radius laser cut template to view the accuracy of the CT-SDR's path.}
		\label{fig:U-shape X-ray}
	\end{figure}

	U-shape drilling is an extension and extreme representation of the J-shape planar drilling concept, in which a NiTi steering guide is inserted and held by the CT-SDR with the cutting plane parallel to the optical table's surface. However, the steering guide's used in U-shape drilling are much longer and take the drill tip through a nearly 180\degree 
	rotation as the drill tip is actuated through a circular trajectory. For this test, a 39.9 mm radius steering guide with a length of 120 mm was used. A 10 PCF sawbone test sample was secured with the front face of the sample perpendicular to the CT-SDR's initial cutting direction. The drill motor was accelerated to 8250 rpm, which in turn rotates the CT-SDR's cutting tip at the same speed. Once the drill tip was at the desired cutting speed, the drill was actuated forward at 1.6 mm/s for the length of the steering guide. When the steering guide reached the end of its length, the drill motor was powered off and the C-arm was used to take X-ray images of the CT-SDR's tip position within the sample, and a laser cut template corresponding to an ideal 35 mm radius steering guide was used to determine the accuracy of the drilled U-shape trajectory. The angle of the final cut was measured from the analyzing the angle between the insertion orientation of the CT-SDR's drill tip and the final orientation. This could be measured via the X-ray images taken of the test. Fig. \ref{fig:U-shape X-ray} shows the X-ray view of the drilled U-shape trajectory and the used laser cut template to evaluate the accuracy of the CT-SDR. Also, Fig. \ref{fig:X-ray Progress} shows the X-ray images demonstrating progression of the CT-SDR through the Sawbone samples during this experiment.
	\subsubsection{Cavity Drilling}\label{cavity}
	
\begin{figure}[!t] 
		\centering 
		\includegraphics[width=1\linewidth]{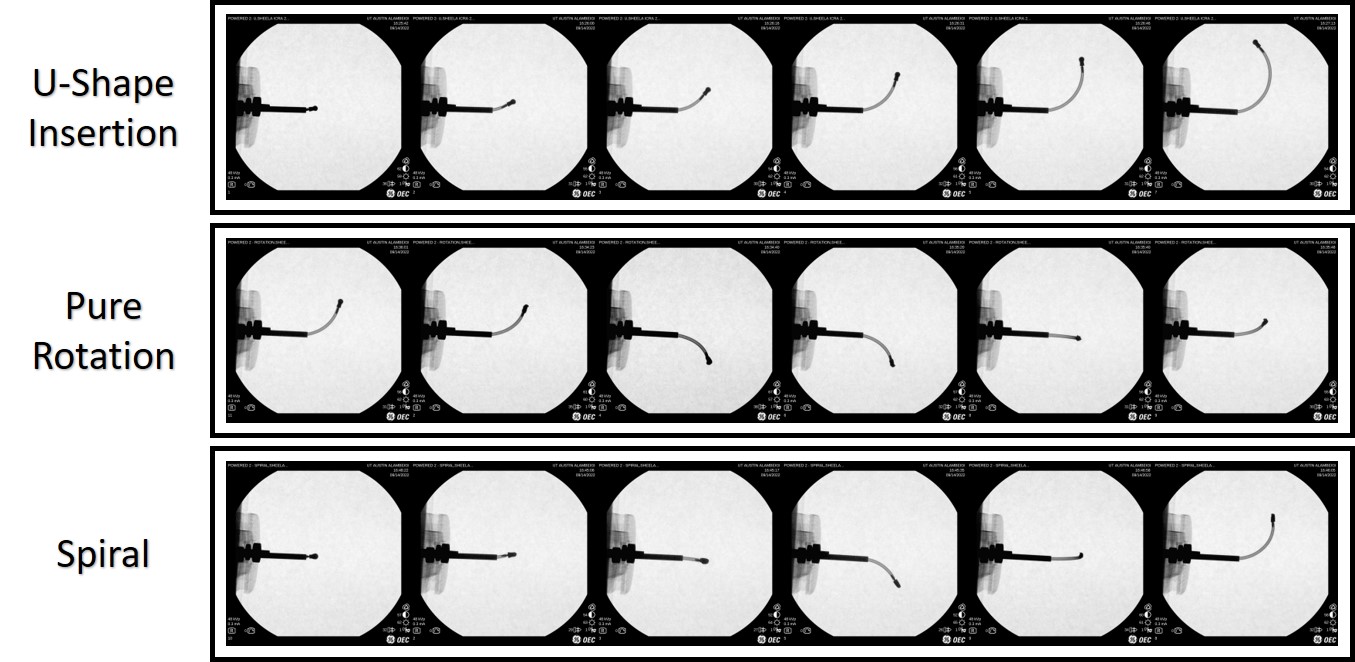}
		\caption{X-ray images showing progression of a test by moving the CT-SDR through free space with a 39.9 mm radius steering guide. Top: U-shape trajectory view. Middle: A singular rotation of a pure rotation test, in which the CT-SDR would do a pure rotation at several depths of cut. Bottom: A spiral test in which the rotation and insertion DoFs move together.}
		\label{fig:X-ray Progress}
\end{figure}

	The original design of the CT-SDR was centered around its ability to produce out-of-plane cuts both through multiple J-shape branch trajectories and through simultaneous rotations and insertions while within a test sample. Several variations of cavity drilling tests were performed with both the 39.9 mm and 71.1 mm steering guides to evaluate the success of the CT-SDR's design. In each test, the sawbone sample was secured in front of the CT-SDR, the drill motor was accelerated to 8250 rpm, and the test was carried out with an insertion speed of 1.6 mm/s and a rotation speed of 9.6\degree /s (unless otherwise specified). Fig. \ref{fig:X-ray Progress} shows the X-ray images demonstrating progression of the CT-SDR through the Sawbone samples during the cavity drilling scenarios.
	
	\textit{(i) J-shape Branch Drilling:} To test the initial functionality of the CT-SDR's rotational DoF, a J-shape branch test was designed. In this test, the NiTi steering guide was actuated through the test sample in a J-shape trajectory, retracted fully, rotated out of plane, and re-inserted through the same entry point to drill another J-shape trajectory. This insertion/retraction/rotation was repeated until 3 paths had been drilled from the same entrance hole. 
	
	\textit{(ii) Independent  2-DoF Drilling (Pure Rotational motion): } In this test, the CT-SDR was tested with both insertion and rotation  DoFs of the steering guides while the CT-SDR's drill tip was within the test specimen. For this test, the CT-SDR was inserted approximately 10 mm and then the insertion was paused as the CT-SDR performed a full rotation of the steering guide before inserting and rotating again. This process was repeated for the length of the steering guide used for the test. 

\textit{(iii) Simultaneous  2-DoF Drilling (Spiral motion): } The final cavity tests conducted with the CT-SDR were centered around coupling the insertion and rotation DoFs together. For this test, the system was set to run with an insertion speed of 0.96 mm/s and a rotation speed of 4.7\degree /s, chosen to ensure that a full rotation would occur before the CT-SDR had translated a full length of the drill tip.

 		\subsubsection{Experimental measurement}
  
		After the conclusion of each test, as the resulting cavities were not clearly visible with a C-arm, 3D models were made  from the drilled test samples. Plaster was poured through the entrance hole of the test and allowed to harden. The Sawbone material was then removed to leave an inverted view of the drilled cavity. These models were then laser scanned (Space Spider, Artec3D) and imported into 3D CAD software (SolidWorks, Dassault Systèmes) where they could be measured and analyzed. Fig. \ref{fig:Laser Model} shows the exemplary plaster and laser scanned models.
	
	

	\section{Results and Discussion}
	\vspace{-5 pt}

	
 Fig. \ref{fig:U-shape X-ray} shows the results of the U-shape drilling with a NiTi steering guide with a radius of curvature of 35 mm. From the figure, it is clear that the CT-SDR can drill around an obstacle and reach a point 82 mm in a perpendicular direction to the entry trajectory. In this experiment, the angle of change in which the CT-SDR's cutting tip has moved through during the test (as measured counter-clockwise from it's original position) was 153\degree. 
	Previous studies that have documented their orientation change throughout their curved drilling process, such as Wang et al. \cite{9732206}, have shown orientation angle changes as large as 65\degree for wrist hinge drilling devices; Alambeigi et al. \cite{alambeigi2017curved} showed a change of up to 40\degree in their experiments. To our knowledge, this is the is the first successful U-shape drilling in a hard tissue.
	
\begin{figure}[!t] 
		\centering 
		\includegraphics[width=0.9\linewidth]{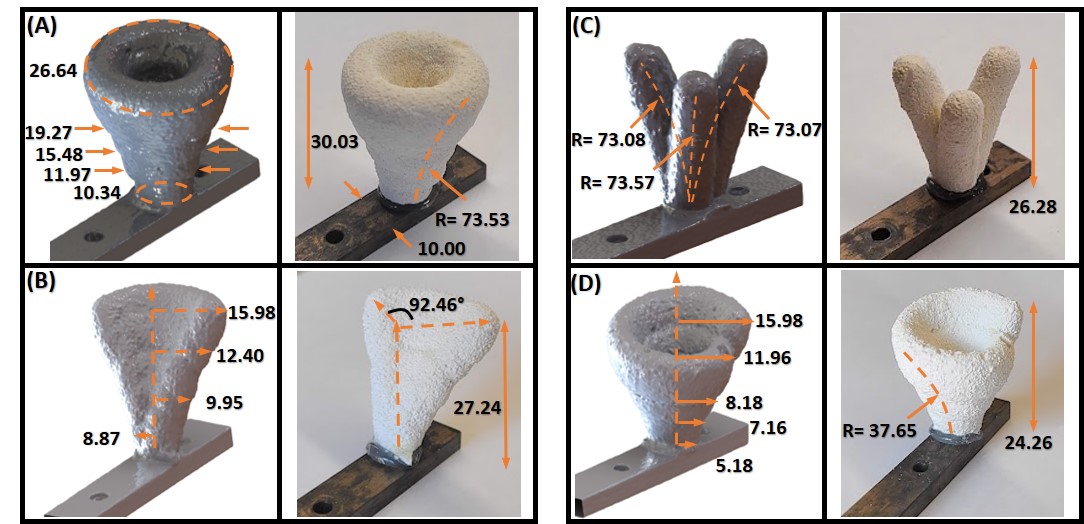}
		\caption{3D renderings (left) of actual cavity drilling models (right). (a) A Pure Rotational test performed in 10 PCF sawbone with the 71.1 mm steering guide. (b) A Pure Rotational test performed in 5 PCF sawbone with the 39.9 mm steering guide, though rotated only 92\degree instead of a full 360\degree. (c) A Branches test in 10 PCF sawbone with the 71.1 mm steering guide. (d) A spiral test performed in 5 PCF sawbone with the 39.9 mm steering guide.}
		\label{fig:Laser Model}
	\end{figure}

	The performed cavity tests lead to the following results:
\textbf{Branches}:
	In the branches tests (Fig. \ref{fig:Laser Model}-C), the goal was to reach locations out of plane with accurate trajectories created by the steering guides. In a block of 5 PCF sawbone, the steering guide with a radius of 71.1 mm produced 3 branches with an average radius only 2.6\% different from the guide, showing that out of plane cuts had little to no effect on the CT-SDR's behavior. 
		
\textbf{Pure Rotational:}
	Perhaps the most significant test run by the CT-SDR, the addition of actuating the rotational DOF during the drilling process created the first true cavities seen by this system. Shown in Fig. \ref{fig:Laser Model}-A, the laser scanned render of the cavity rotation test run in 10 PCF sawbone with the 71.7 mm steering guide. From the first rotation within the material to the final ring made by the drill tip, the diameter of the cut material more than doubled, going from 10.34 mm to 26.64 mm in a distance of 30.03 mm. The rotational tests took approximately 2min, significantly faster than Alambeigi et al. \cite{alambeigi2017curved} in which each curved path took up to 5-9 min. Tests were also run with a steering guide of radius 39.9 mm, instead of rotating a full 360\degree at each insertion step, the CT-SDR was rotated 92.46\degree to create a partial cavity in 5 PCF sawbone shown in Fig. \ref{fig:Laser Model}-B. The diameter of the projected cut increased from 17.74 mm to 31.96 mm. 
	
	
\textbf{Spiral:}
	The coupled insertion/rotation cavity tests displayed similar success to their decoupled counterparts. Drilling through 5 PCF sawbone, the 39.9 mm radii steering guide was used while simultaneously actuating the rotational and translational components together. The result was a cavity similar in size and dimension to those in previous tests, with an insertion radius of 5.18 mm and a final radius of 15.98 mm as measured as the distance from the final drilling location to the axis of rotation. However, instead of rings that increase in diameter as the test progressed, a spiral shape was formed. Comparing between the decoupled and coupled DoF tests, decoupled pure rotation had better accuracy on holding to the trajectory of the steering guide, only 3.4\% away from the desired 71.1 mm radius. These comparisons can be seen between the different laser scanned models in Fig. \ref{fig:Laser Model}, along with the smooth surface quality.

\begin{figure}[!t] 
		\centering 
		\includegraphics[width=0.8\linewidth]{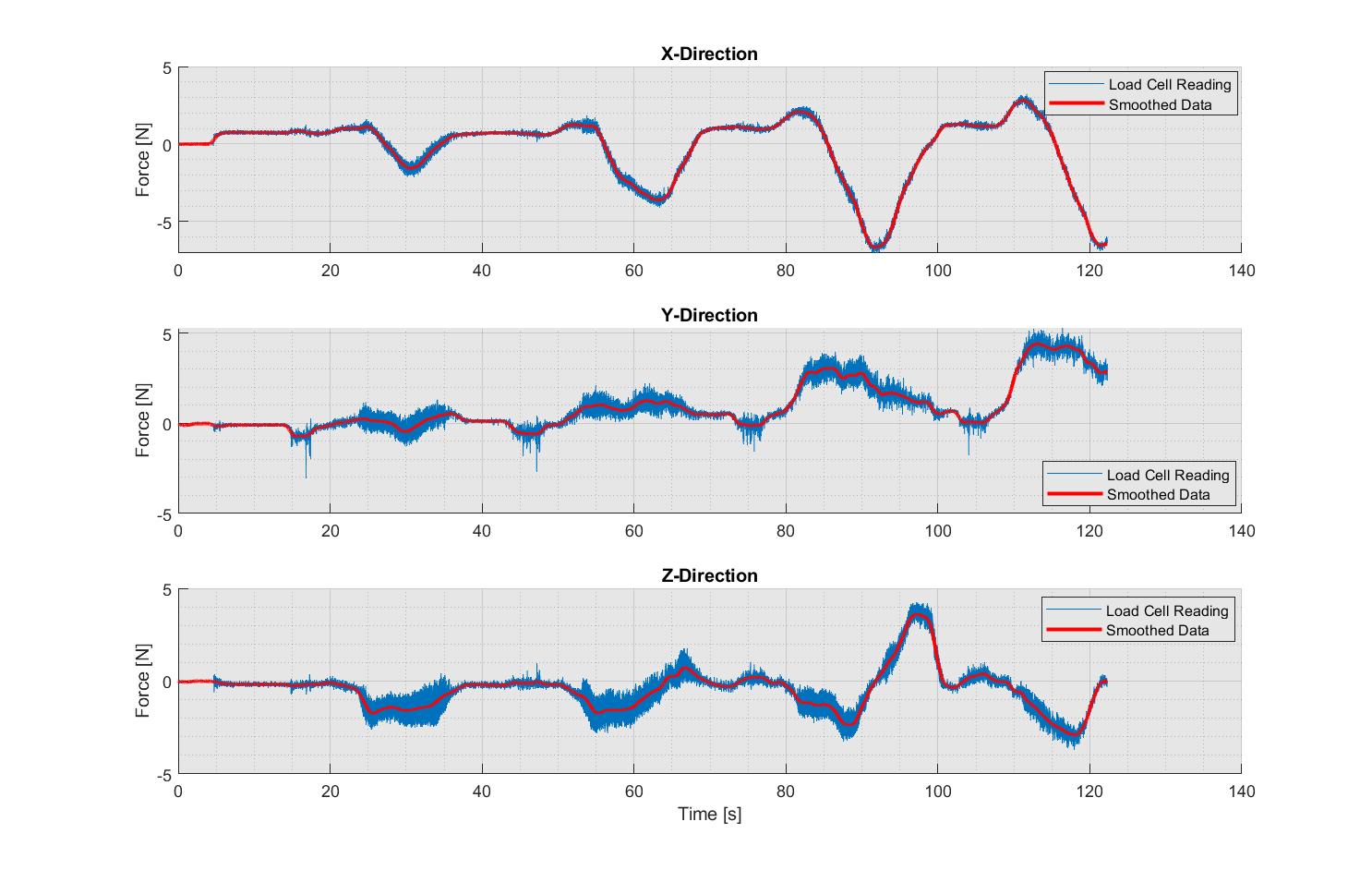}
		\caption{Components of both the measured and smooth forces during a pure rotational cavity drilling test performed in 10 PCF sawbone with the 71.1 mm steering guide.}
		\label{fig:Cavity Forces}
	\end{figure}

	Fig. \ref{fig:Cavity Forces} illustrates the average directional loads applied to the drill tip and the test sample during a pure rotational cavity drilling experiment using a 10 PCF sawbones block and 71.1 mm radius steering guide. The forces were captured by the load/torque cell (with frequency of $1 kHz$) and were smoothed with a span of 100 and averaged in MATLAB using the \textit{smooth} function (MATLAB, MathWorks). The forces felt in both the X and Z-directions oscillate as the test progresses, as these directions are both perpendicular to the direction of initial cut. As the drill bit cycles around, it's primary cutting force is rotated in different directions. The force felt in the Y-direction increased each time the drill was inserted further into the test sample. When the components of the drilling force are resolved into a singular magnitude, the maximum force felt throughout the experiment was 7.13 N. While this magnitude is slightly larger than other drills in literature (e.g., \cite{alambeigi2017curved, 9732206, alambeigi2016toward}) this could be attributed to the speed at which the drill tip was moving at the time of the end of the test. 
		\section{Conclusion and Future Work}
	\vspace{-4 pt}
	With the goal of improving current minimally invasive spinal tumor treatments, we proposed a novel two DoF CT-SDR system to address a lack of dexterity seen in  the existing rigid surgical instruments. The proposed design was evaluated for both direct path and cavity drilling scenarios, and proved itself capable in both regards. Reaching up to 82 mm in the perpendicular direction to the point of entry, and surpassing 150\degree angles with the drill tip for U-shaped path drilling was one of the unique features of the proposed robotic system. Moreover, the performance of the system was verified in accurate out-of-the plane J-shape branch and cavity cutting scenarios; performing tests in approximately 2 min, faster than previous studies \cite{alambeigi2017curved}.
	
		In the future, we will take advantage of the modular design of the proposed CT-SDR  by adding additional carriages for enabling long S-shape cutting trajectories. We will also evaluate our system on an integrated robot-assisted steerable drilling procedure on animal bones and human cadaveric specimens \cite{sefati2020surgical, wilkening2017development,alambeigi2018convex}. 
	
\bibliographystyle{IEEEtran}
\bibliography{ISMR2023_Handheld}

	\end{document}